\RequirePackage{tikz}
\usetikzlibrary{arrows}
\usetikzlibrary{shapes.geometric}
\usetikzlibrary{shapes.misc, positioning}
\usetikzlibrary{shapes.multipart}
\usetikzlibrary{arrows,decorations.pathreplacing}
\documentclass[sn-mathphys,pdflatex,iicol]{sn-jnl}

\usepackage{xcolor}
\usepackage{placeins}
\usepackage{natbib}
\usepackage{gensymb}
\usepackage{subcaption}
\usetikzlibrary{positioning}
\usetikzlibrary{matrix}

\usepackage{microtype}
\usepackage{xspace}
\newcommand{\eg}{e.\,g.,\xspace}
\newcommand{\ie}{i.\,e.,\xspace}

\jyear{2022}%

\theoremstyle{thmstyleone}%
\theoremstyle{thmstyletwo}%

\theoremstyle{thmstylethree}%

\begin{document}

\title[~]{Graph Neural Networks for Multivariate Time Series Regression with Application to Seismic Data}

\author*[1,2]{\fnm{Stefan} \sur{Bloemheuvel}}\email{s.d.bloemheuvel@jads.nl}
\equalcont{These authors contributed equally to this work.}

\author[1,2]{\fnm{Jurgen} \sur{van den Hoogen}}\email{j.o.d.hoogen@jads.nl}
\equalcont{These authors contributed equally to this work.}

\author[3,4,7]{\mbox{\fnm{Dario} \sur{Jozinovi\'{c} }}}\email{djozinovi@gmail.com}

\author[3]{\mbox{\fnm{Alberto} \sur{Michelini}}}\email{alberto.michelini@ingv.it}

\author[5,6]{\mbox{\fnm{Martin} \sur{Atzmueller}}}\email{martin.atzmueller@uni-osnabrueck.de}

\affil*[1]{\orgname{Tilburg University}, \orgaddress{\city{Tilburg}, \country{The Netherlands}}}

\affil[2]{\orgname{Jheronimus Academy of Data Science}, \orgaddress{\city{'s-Hertogenbosch}, \country{The Netherlands}}}

\affil[3]{Istituto Nazionale di Geofisica e Vulcanologia, Rome, Italy}

\affil[4]{Department of Science, Università degli Studi Roma Tre, Rome, Italy}

\affil[5]{\orgname{Osnabr\"uck University}, \orgaddress{Semantic Information Systems Group, \city{Osnabr\"uck}, \country{Germany}}}

\affil[6]{\orgname{German Research Center for Artificial Intelligence (DFKI)}, \city{Osnabrück},  \country{Germany}}

\affil[7]{now at Swiss Seismological Service (SED) at ETH Zurich, Zurich, Switzerland}

\abstract{Machine learning, with its advances in deep learning has shown great potential in analyzing time series. In many scenarios, however, additional information that can potentially improve the predictions is available. 
This is crucial for data that arise from \eg sensor networks that contain information about sensor locations. Then, such spatial information can be exploited by modeling it via graph structures, along with the sequential (time series) information.

Recent advances in adapting deep learning to graphs have shown potential in various tasks.
However, these methods have not been adapted for time series tasks to a great extent.
Most attempts have essentially consolidated around time series forecasting with small sequence lengths. 
Generally, these architectures are not well suited for regression or classification tasks where the value to be predicted is not strictly depending on the most recent values, but rather on the whole length of the time series. 

We propose TISER-GCN, a novel graph neural network architecture for processing, in particular, these long time series in a multivariate regression task.
Our proposed model is tested on two seismic datasets containing earthquake waveforms, where the goal is to predict maximum intensity measurements of ground shaking at each seismic station.

Our findings demonstrate promising results of our approach -- with an average MSE reduction of 16.3\% -- compared to the best performing baselines. In addition, our approach matches the baseline scores by needing only half the input size. The results are discussed in depth with an additional ablation study.
}


\maketitle

\section{Introduction}\label{sec1}

In today's world, advances in hardware and wireless network technology have opened the path for energy-efficient, multifunctional and low-cost sensors~\cite{tilak2002taxonomy}. Spread across a large geographical region, a set of sensors can then form a sensor network used for data collection and analysis~\cite{tubaishat2003sensor}, in particular considering large-scale time series data. Example domains where such real-world sensor data is analyzed include, \eg traffic \cite{aslam2012city}, weather \cite{hatchett2020observations} and seismology \cite{van2020automated}, in particular regarding time series regression and classification, \eg~\cite{chen2016xgboost,tan2021time}.

Recently, there have been considerable advances in deep learning methods, in particular regarding CNNs, with respect to their ability to automatically find structure and meaningful features in the data. This leads to powerful (implicit) feature construction and computationally efficient models, \eg~\cite{hoogen2020improvedWDCNN,ince2016real} for time series.
However, if only the time series data are examined, then some aspects of the sensor data are left unseen, \ie the spatial relations of sensors in datasets that are geographically grounded.

Consequently, researchers have developed deep learning techniques to perform time series analysis like forecasting \cite{wu2020connecting}, anomaly detection \cite{deng2021graph} and imputation \cite{cini2022filling},  with data arising from networks (\ie graphs), called graph neural networks (now referred to as \emph{GNNs}), which we also focus on in this paper.
However, if the predicted value does not rely more on recent values from the input than early values (known as Time Series Extrinsic Regression (TSER) \cite{tan2021time}), the aforementioned models are not adequate for the task.

Therefore, in this paper we tackle the problem of multivariate time series regression, for which we present a novel GNN-based architecture named TISER-GCN. Our evaluation applies high-frequency network-based seismic data demonstrating the efficacy of our proposed approach.

Previous attempts for tackling similar time series problems with graph-based methods have been made by \cite{van2020automated,yano2021graph,kim2021graph}, yet each has some shortcomings. 
van den Ende and Ampuero \cite{van2020automated} mention that they designed a GNN for the localization of earthquakes from waveform data. 
However, they only append the (latitude, longitude) information to the time series being handled by a CNN. 
Therefore, while prediction scores improved, no actual GNN layers were used.
Second, \cite{yano2021graph} proposed a graph partitioning algorithm that works together with a CNN. However, they make use of classical graph theory techniques and a GNN method is not applied.
Lastly, \cite{kim2021graph} recently suggested a method that uses CNNs and GNNs for seismic event classification. 
However, (1) no spatial information is used at all, \ie each edge has a weight of 1,  nor (2) meta information about the stations is added, and (3) only three nodes are examined for each observation, which could be difficult to interpret as a full-fledged/complex network.

Therefore, we propose a larger scale GNN architecture that can process multivariate time series for such a regression task. 
By combining the capabilities of convolutional layers (feature extraction) and graph convolutional layers (spatial information), our model can manage the feature sizes that are common in high-frequency time series data arising from multiple sensors. 

We test our proposed model on network-based seismic data, which serve as an intuitive domain where GNN models could be operated due to the naturally geographical-grounded sensors. The model is inspired by the work presented in \cite{jozinovic2020rapid} and  \cite{jozinovic2021transfer}, which functions as our most prominent baseline. In their work, the maximum ground-shaking at a set of seismic stations is predicted by tackling this as a regression problem. Their model used convolutional layers to extract useful features from a given time series. We start from their work as a departure point, and illustrate how to design the deep learning model structure using GNNs for such a task. 

\noindent Our contributions are summarized as follows:
\begin{enumerate}
\item We propose a method to perform multivariate regression on time series originating from graph-structured data. For this, we present an architecture utilizing convolutional and graph convolutional layers that is also adjustable for other use cases or datasets, \eg time series classification tasks.
\item We evaluate our model thoroughly on two seismological datasets that differ significantly from one another evidencing the generality and potential of the proposed GNN-based architecture in this task. We discuss our results in detail and perform a comparison against several baseline models (in particular \cite{jozinovic2020rapid}, but also \cite{kim2021graph,velickovic2018graph}) and traditional machine learning methods.
\item Finally, we systematically analyze the capabilities of our model in detail by comprehensive experimentation adjusting several hyperparameters in our proposed workflow.
\end{enumerate}

The article is further structured as follows: we discuss related work in Sect.~\ref{sec:relatedwork}, which provides the necessary background on deep learning, graphs and GNNs. 
Next, Sect.~\ref{sec:method} introduces the dataset, our method and training settings.
After that, Sect.~\ref{sec:results} presents our results and discusses these in the context of a model-based comparison.
Finally, Sect.~\ref{sec:conclusions} concludes with a summary and outlines interesting directions for future work.

\section{Background and related work}\label{sec:relatedwork}

This section briefly outlines the background and related work on graphs and deep learning in general, CNNs, GNNs and its utilization in time series, as well as the implementation of deep learning for seismic analysis.

\subsection{Deep learning on complex data}
Traditional machine learning often requires considerable effort from the user to construct meaningful features, which usually is rather time-consuming and error-prone \cite{hoogen2020improvedWDCNN}.
Deep Learning provides a way for automatic feature extraction with help of multiple layers that can utilise nonlinear processing.
In particular, this also relates to complex representations such as multivariate time series and graphs.
Therefore, deep learning offers strong processing and learning on complex data.

Initially, the multilayer perceptron (MLP) was developed in which all network layers are fully linked \cite{rumelhart1986learning}. 
While being powerful, due to its high computation time the depth of the network is limited.
Therefore, researchers have found ways to create more advanced architectures for specific tasks. 
One of the most prominent and successful outcomes of this effort is the CNN.

A CNN is a regularized MLP that is specialized in handling data structures with multiple dimensions (\eg pictures with color channels). 
It uses a feed-forward structure with convolutions instead of more general matrix multiplications. 
CNNs have been widely adopted in natural language processing and Computer Vision. 
A CNN has an advantage over MLPs due to its use of weight sharing, sampling and local receptive fields~\cite{goodfellow2016deep}. 

For creating output, the convolutional layers convolve the input using filters and activation functions. 
A convolution operation is defined as

\begin{equation}
    y_{i}^{l+1}(j)=k_{i}^{l} \cdot M^{l}(j)+b_{i}^{l},
\end{equation}

\noindent where $y_{i}^{l+1}(j)$ denotes the input of the $j$-th neuron in the feature map $i$ of layer $l+1$, $k_{i}^{l}$ the weights of the $i$-th filter kernel in layer $l$, $M^{l}(j)$ the $j$-th local region in layer $l$ and $b_{i}^{l}$ the respective bias.
An activation function is applied after each convolutional layer to retrieve the nonlinear features.

\subsection{Graphs}
\begin{figure}[htb]
    \begin{minipage}[t]{.45\textwidth}
        \centering
        \begin{tikzpicture}
            \matrix  [{matrix of math nodes},left delimiter={[},right delimiter= {]}] (m)
            {
            2 & 7 & 1 & 4 & 3 & 0 & 0 & 1 & 2 \\
            5 & 9 & 6 & 2 & 9 & 1 & 3 & 0 & 3 \\
            7 & 8 & 1 & 5 & 7 & 6 & 6 & 2 & 7 \\
            0 & 2 & 3 & 0 & 1 & 5 & 8 & 7 & 2 \\
            1 & 5 & 7 & 1 & 4 & 1 & 9 & 5 & 1 \\
            };  
            \draw[color=red] (m-1-1.north west) -- (m-1-3.north east) -- (m-3-3.south east) -- (m-3-1.south west) -- (m-1-1.north west);
            \draw[color=blue] (m-2-6.north west) -- (m-2-8.north east) -- (m-4-8.south east)  -- (m-4-6.south west) -- (m-2-6.north west);
            \end{tikzpicture}
        \subcaption{}\label{fig:convolutiononmatrix}
    \end{minipage} 
    \hfill
    \begin{minipage}[t]{.45\textwidth}
        \centering
        \begin{tikzpicture}[
    roundnode/.style={circle,draw=black!50, very thick, minimum size=5mm},
    squarednode/.style={rectangle, draw=red!50, very thick, minimum size=20mm},
    squarednode2/.style={rectangle, draw=blue!50, very thick, minimum size=20mm},
    ]
        \node[roundnode] at (0,0) (0) {0};
        \node[roundnode] at (-1,1.15) (1) {1};
        \node[roundnode] at (-2.5,0.4) (2) {2};
        \node[roundnode] at (-3.4,-0.3) (3) {3};
        \node[roundnode] at (-1.2,-0.6) (4) {4};
        \node[roundnode] at (-2.4,-0.8) (5) {5};
        
        \node[squarednode2] at (-0.6,-0.35) (6) {};
        \node[squarednode] at (-2.9,-0.2) (7) {};

        \draw[] (0) -- (1);
        \draw[] (3) -- (5);
        \draw[] (2) -- (1);
        \draw[] (4) -- (0);
        \draw[] (3) -- (2);
        \draw[] (3) -- (5);
        \draw[] (2) -- (4);
        \draw[] (2) -- (5);
        
    \end{tikzpicture}
        \subcaption{}\label{fig:convolutionongraph}
    \end{minipage} 
    \caption{Examples: (a) Matrix-based convolution on an image / time series or (b) a graph, for which a convolution is a lot harder to define than in (a).}
    \label{fig:convolutionprocedure}
\end{figure}
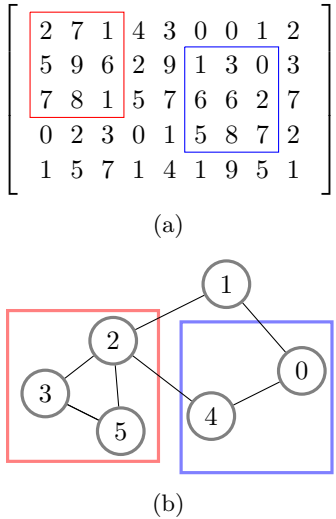

Before discussing the extension of deep learning models to graphs, we first introduce some background and basic notation. 
We define a graph $G$ as $G=(V,E)$ where $V$ is the set of nodes and $E$ the set of edges (see Fig.~\ref{fig:convolutionongraph} for an example).
An edge $e_{ij}=(v_i,v_j)$ connects two nodes $v_{i}, v_{j} \in V$.
A common way to represent a graph is with an adjacency matrix $A \in \mathbb{R}^{N \times N}$ where $N=\lvert V \rvert$, which is a square matrix such $A_{ij}=1$ if there is an edge from node $v_{i}$ to node $v_{j}$, and 0 otherwise.
The number of neighbors of a node $v$ is known as the degree of $v$ and is denoted by $D_{ii} = \sum\nolimits_{j}^{}A_{ij}$, where $D$ is then the diagonal degree matrix.
Edges can be undirected and directed. Undirected edges contain no notion of source and destination, \eg the absolute distance between two nodes is always equal no matter from which node the measurement starts. 
Directed edges do contain direction information, \eg whether somebody follows someone else on a social network or not.

In addition, nodes and edges (as well as entire graphs) can have \emph{features} as well, such that a feature vector $a=(a_1, a_2, \ldots, a_n)$ of individual features \mbox{$a_i \in \Omega_A$} out of a feature domain $\Omega_A$ is assigned to the nodes (and/or edges). 
GNN problems therefore mostly consist of node-level, edge-level and graph-level tasks, utilizing the aforementioned feature types. 

However, standard convolutional layers (\eg CNNs) are not applicable to graph-structured data due to its non-euclidean nature.
In particular, one cannot convolve an $n \times n$ grid over a graph the same way as with an image.
Fig. \ref{fig:convolutionprocedure} shows an example: both the red and blue boxes convolve over the same grid with $3 \times 3$ numbers (Fig. \ref{fig:convolutiononmatrix}).
The red box convolves three nodes, while the blue box convolves over two nodes as shown in Fig. \ref{fig:convolutionongraph}.
Thus, extensive effort was put into finding ways to define convolutions over graphs.

\subsection{Graph Neural Networks}
GNNs are deep learning based methods that are adapted for the graph domain.
In general, the history of creating deep learning models for graphs is surprisingly long.
For example, Recursive Neural Networks were already adapted to work on directed acyclic graphs in the 1990s~\cite{sperduti1997supervised}.
However, one recent paper revamped the interest in using deep learning on graphs \cite{bruna2013spectral}.
They propose two ways that use hierarchical clustering and the spectrum of the graph Laplacian to perform convolutions on low-dimensional graphs.
The approach from \cite{bruna2013spectral} falls into one of the two historical main methods to perform convolutions with graphs: (1) Spectral methods and (2) Spatial methods. 

Spectral methods use the eigenvectors and eigenvalues of a matrix with eigendecomposition, and perform convolutions using the Graph Fourier Transformation and the inverse Graph Fourier transform, respectively. 
These transformations of the signal $x$ are defined as $F(x)=U^{T}x$ and $F^{-1}(x)=Ux$, where $U$ represents the matrix of eigenvectors of the normalized graph Laplacian $L=I - D^{-1/2}AD^{-1/2}$, $D$ is the degree matrix of the adjacency matrix $A$ and $I$ refers to the identity matrix of length $\lvert V \rvert$~\cite{zhou2020graph}.

Spatial methods use \emph{message passing} techniques, which consider the local neighborhood of nodes and perform calculations on their top-k neighbors.
With a node aggregation/update function $f$, an updated node representation $Z$ could then be defined as $Z=f(G)X$ where $G$ refers to the adjacency or Laplacian matrix, and $X$ to the node features of the nodes contained in $G$ \cite{chen2020bridging}.
However, a serious issue with spatial methods is in determining the convolution procedure with differently sized node neighborhoods \cite{zhou2020graph}.

To conclude, there are two typical operations when designing GNNs: Spatial methods focus more on the connectivity of the graph, while Spectral methods focus on its eigenvalues and eigenvectors~\cite{chen2020bridging}. 
Both approaches were then simplified by Kipf and Welling~\cite{kipf2016semi} into the so-called Graph Convolutional Networks (GCNs), which are also used in this paper.
They define their propagation rule (convolution in a graph) as follows:

\begin{equation}
    H^{(l+1)}=\sigma\left ( \tilde{D}^{-\frac{1}{2}}\tilde{A} \tilde{D}^{-\frac{1}{2}}H^{(l)}W^{(l)} \right )
\end{equation}

\noindent where $H^{(l)}\in \mathbb{R}^{N\times D}$ is the matrix of activations of the $l$th layer, $\sigma$ denotes the selected activation function, $\tilde{D}={\textstyle\sum}_{j}^{}\tilde{A}_{ij}$ refers to the degree matrix; matrix $\tilde{A}=A+I_{N}$ is the adjacency matrix of the undirected graph $G$ with the added self-connections $I_{N}$ to include a node's own node features, $H^{(0)}=X$ where $X$ are the node features and $W^{(l)}$ is the trainable weight matrix for a specific layer. 

A different method was proposed by \cite{velickovic2018graph} (graph attention networks, now referred to as GAT),
where the structural information of $A$ is dropped and is more implicitly defined by using self-attention over the node features. The authors motivate this by referencing previous work (\eg transformers \cite{vaswani2017attention}) that showed that self-attention is sufficient.
Still, both techniques (GCNs and GATs) can produce node-specific outputs of $N \times F$ features, where $F$ is the number of desired output features for each node $N$. Based on this, we will discuss extensions for time series analysis below.

\subsection{Graph Neural Networks for Time Series Analysis}
Considering the connection between GNNs and classical time series analysis, most effort is visible in time series forecasting \cite{cao2020spectral,wu2020connecting}. 
These approaches adapt existing neural network architectures to use operators from the graph domain.
Examples are gated recurrent GNNs that utilise the spectral convolutions from \cite{defferrard2016convolutional}. 
Also, Diffusion-Convolutional Networks are introduced that take into account the in and out-degree of nodes to capture the spatial dependencies of nodes better, which is beneficial in \eg traffic prediction \cite{li2017diffusion}. 
Later on, spatio-temporal graph convolutional neural networks are introduced that interchange the convolution procedure between the temporal and spatial dimensions~\cite{yu2017spatio}. 
In addition, GNNs have been used to perform anomaly detection in time series data. 
\cite{deng2021graph} propose an attention-based GNN that used the results of a forecast to classify deviating predictions as anomalies. 
In addition, \cite{cini2022filling} propose GRIL (Graph Recurrent Imputation Layer), a Spatial-Temporal GNN that reconstructs missing data by learning spatial-temporal representations.

To conclude, a lot of progress has been made in combining GNNs with classical time series related tasks. 
However, time series regression (and classification) tasks have not received the same amount of attention yet.
Especially when the target value does not rely on more recent values from the input, but rather on the whole length of the time series, other model architectures are needed.
As discussed below, seismic data is a typical domain where these data characteristics naturally arise.

\subsection{Deep Learning for Seismic Analysis}
Over the past decades, huge volumes of continuous seismic data have been collected
\cite{ingate_iris_2008, strollo_eida_2021}. 
With the availability of large datasets and advances in machine learning, the seismological community has also seen a rise in the use of machine and deep learning. Exemplary use cases are magnitude estimation \cite{ochoa2018fast} and earthquake detection \cite{mousavi2020earthquake}.
Specifically for waveform analysis, the CNN has been applied several times. 
For example, \cite{lomax2019investigation} developed a CNN for single-station localization, magnitude and depth estimation. 
In addition, CNNs were developed for P- and S-wave arrival times picking \cite{ross2018p, mousavi2020earthquake}.
Others used multi-station waveforms which were analysed for estimating the earthquakes' location~\cite{van2020automated,kriegerowski2019deep} or early warning~\cite{munchmeyer2021transformer}.

The work of \cite{van2020automated,yano2021graph,kim2021graph} comes most close to the goals of our work. 
Each of these papers tried to use time series related data in combination with graphs to improve predictions. 
\cite{van2020automated} used classical CNNs that attached the (latitude, longitude) locations of the sensors to the waveforms to improve predictions, which differs from our goal to use GNN layers.
In other words, metadata was used to enhance their CNN model with spatial-information, except no graph layers were applied.
\cite{yano2021graph} propose a technique that combines CNNs with graph partitioning to group time series together based on spatial information. 
This procedure increases the quality of the within-group features, and improves predictions, but no GNNs were utilised.
Another recently proposed method for handling seismic data with GNNs is from \cite{mcbrearty2022earthquake}. Here, the location and magnitude of earthquakes is predicted.
However, their input are pre-calculated characteristics of the earthquakes for each station, which differs from our goal to use raw waveform data.
Lastly, \cite{kim2021graph} present a method that uses CNNs and GNNs for seismic event classification. 
However, no spatial information is provided to the model (\ie adjacency matrices only containing 1's are created), no meta information of the nodes is added, only three stations are examined simultaneously, and their method focuses on time series classification instead of regression.

In this paper, we specifically propose a technique that will use the full power of GNNs to perform time series analysis. 
To the best of our knowledge, no GNN-based method was used to perform such a time series regression task before.

\section{Method}\label{sec:method}
In this section, we first introduce the datasets used in the experiments. With this as context information,  we then define our problem formally. After that, we describe how to generate networks given the datasets. Next, we present the framework of our proposed model for multivariate time series regression (TISER-GCN). We first provide an intuition on an abstracted level, followed by a detailed discussion of the full architecture shown in Fig.~\ref{fig:modelarchitecture}, and its implementation. Finally, we discuss model training and the applied baseline models which are used for our evaluation.

\subsection{Dataset}\label{sec:dataset}
We perform regression on two datasets recorded by the Italian national seismic network \cite{michelini_italian_2016, danecek_italian_2021}, described fully in  \cite{jozinovic2020rapid, jozinovic2021transfer}. 
GNNs are an ideal candidate for the analysis of seismic data, since seismic measurements contain (1) an enormous amount of data and (2) sensors that are geographically grounded.
Each sensor (\ie seismometer or accelerometers) in the dataset continuously records the amplitudes of the seismic waves resulting from earthquake occurrences along three components of ground motion (i.e., 3 dimensions): up-down, north-south, and east-west. 
The input maximum (i.e., the greatest amplitude detected across all stations during the time window) is used to normalize the data, as performed by \cite{jozinovic2020rapid}.
The data recorded by the sensors located at the stations are crucial for seismologists to understand the nature of the recorded earthquakes  (e.g., magnitude, location, focal mechanism, etc).

Because information (via telecommunication) can be transmitted faster than the seismic waves travel, seismologists have developed algorithms to predict the maximum intensity measurements (IMs) of ground shaking at a set of far-away stations, caused by an earthquake, using only the very first stations that recorded the earthquake already.
In the seismological literature, this objective is known as ``earthquake early warning''.
The IMs used here include \emph{peak ground acceleration} (PGA), \emph{peak ground velocity} (PGV) and \emph{spectral acceleration} (SA) at 0.3, 1 and 3 second periods and represent the labeled data of our model.

Therefore, the task with these datasets is as follows: by using the earthquake recordings from the stations nearby the epicenter, recorded within 10 s from the origin time of the earthquake, we make predictions of the IMs at all stations within the network. 
A large majority of the stations have not yet recorded the maximum earthquake-related ground motion or ground motion at all.
Therefore, we hypothesize that GNNs are highly suited for this time series regression task to predict IMs.

An example of an earthquake (red solid star) drawn from our dataset is shown in the left of Fig.~\ref{fig:regressionnew}. 
After the initial earthquake waves start to spread, only one station (called FEMA) has started recording part of the earthquake, while the other stations farther to the northwest have not recorded the first waves.

The CI dataset consists of 915 earthquakes recorded on a set of 39 stations (CI network) in central Italy.
The earthquake epicenters and station locations are within the area that consists of latitude [42$\degree$, 42.75$\degree$] and longitude [12.3$\degree$, 14$\degree$], with earthquakes happening from 01/01/2016 until 29/11/2016. It contains many spatially concentrated earthquakes and a dense network of stations.
Earthquakes have a depth between 1.6~km $\leq$ z $\leq$ 28.9~km and magnitudes in the range 2.9 $<$ M $\leq$ 6.5. 

The CW dataset consists of 266 earthquakes recorded on a set of 39 other stations (CW network) in central-western Italy. The earthquake epicenters and station locations are within the area bounded by latitudes [41.13$\degree$, 46.13$\degree$] and longitudes [8.5$\degree$, 13.1$\degree$], with earthquakes spanning the time period between 01/01/2013 and 20/11/2017. 
All the earthquakes are in the depth between 3.3~km and 64.7~km, with magnitudes in the range 2.9 $<$ M $\leq$ 5.1.
Therefore, the CW dataset clearly covers a larger area than the CI dataset and, as Fig.~\ref{fig:networkxplots} illustrates, the earthquakes of the CW dataset are scattered across a large part of central and northern Italy whereas the CI dataset has earthquakes concentrated in one small area. 

\tikzset{
state/.style={
       rectangle split,
       rectangle split parts=2,
       rectangle split part fill={blue!20,red!20},
       rounded corners,
       rectangle split horizontal,
       draw=black, thick,
       minimum height=14em,
       text width=1.2cm,
       minimum width=1.5em,
       inner sep=2pt,
       text centered,
       fill opacity=0.5
       }
}

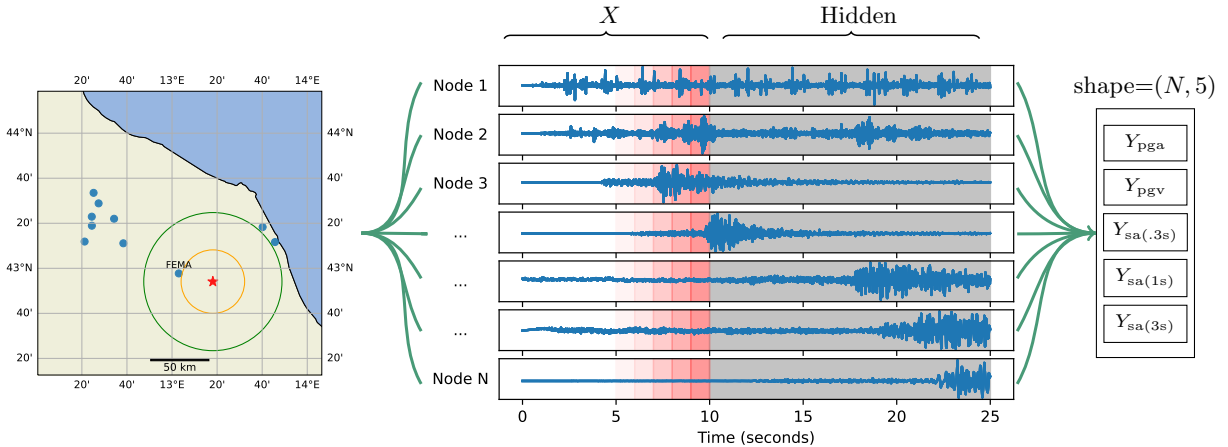
\begin{figure*}[htb]
    \centering
    \begin{tikzpicture}
            \node[inner sep=0pt] (forecast) at (0,0)
                {\includegraphics[width=.5\textwidth]{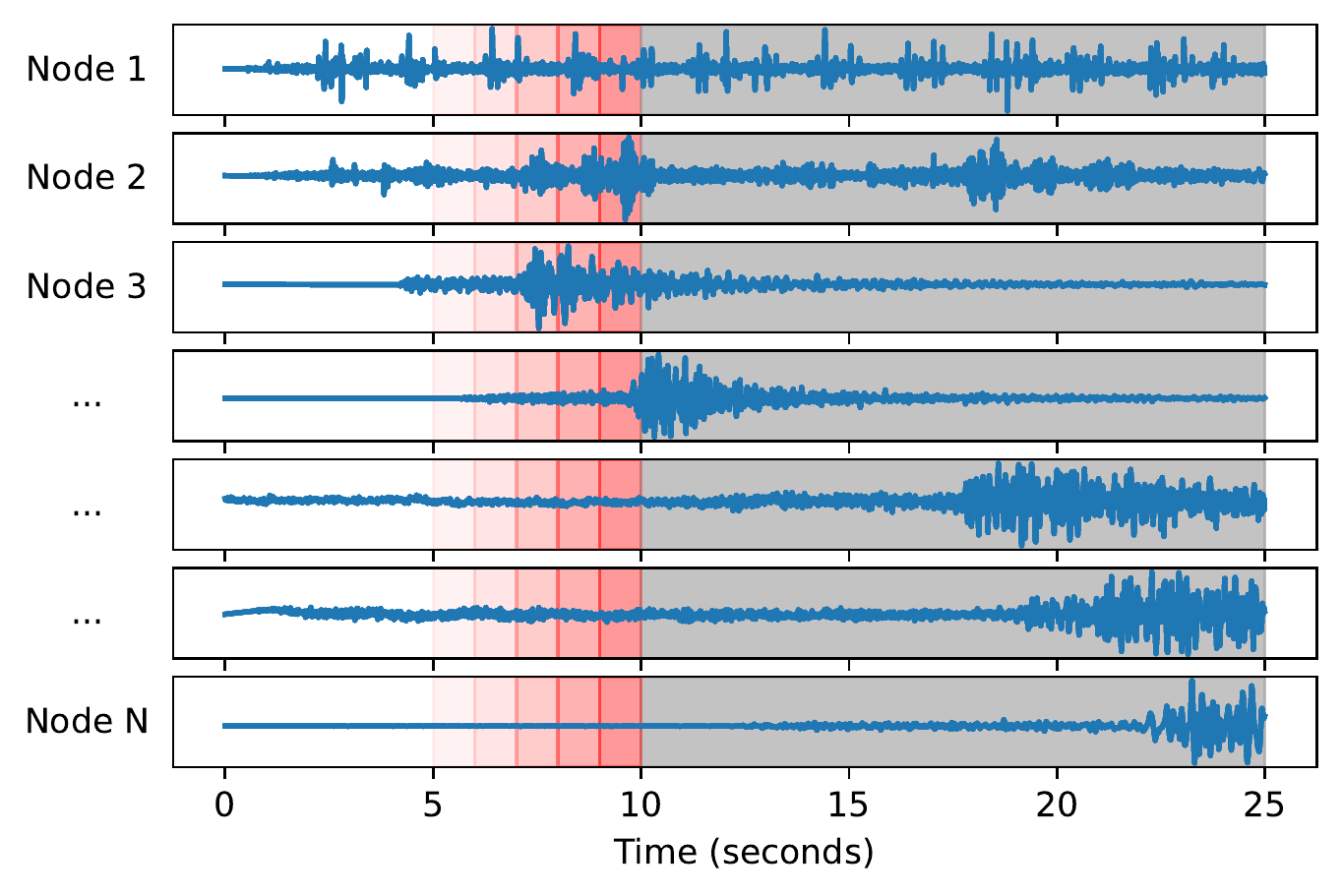}};
            \node[rectangle,draw, minimum height=3.3cm, minimum width=1.3cm] (box) at (5.6,0.3) {}; 

            \node[inner sep=0pt] (earthquakeplot) at (-7.2,0.3)
                {\includegraphics[width=.3\textwidth]{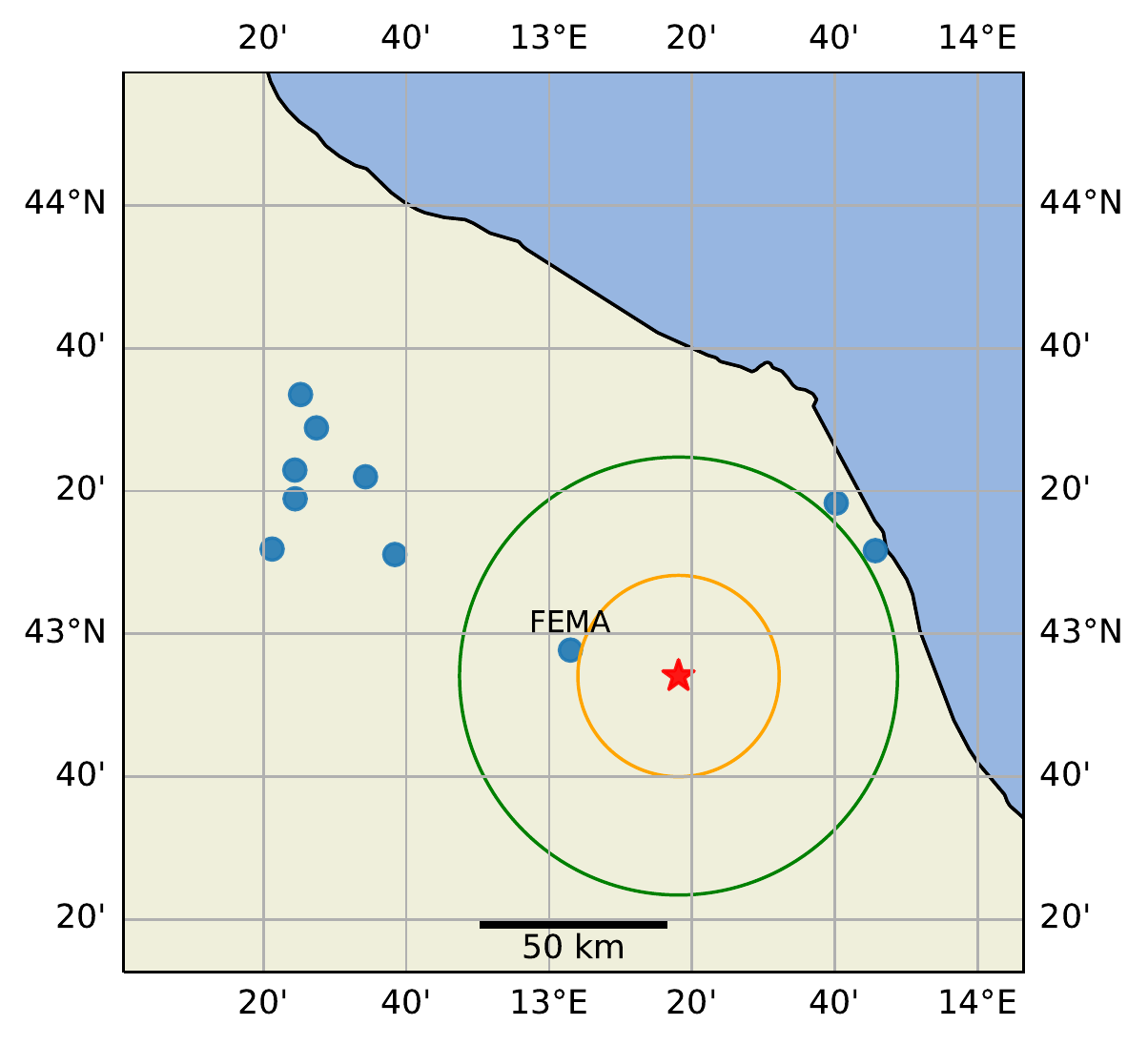}};
            
            \draw [-, color={rgb,255:red,72; green,154; blue,119}, line width=0.4mm] (earthquakeplot.east) to [out=0,in=240] (-4,2.3);
            \draw [-, color={rgb,255:red,72; green,154; blue,119}, line width=0.4mm] (earthquakeplot.east) to [out=0,in=240] (-4,1.6);
            \draw [-, color={rgb,255:red,72; green,154; blue,119}, line width=0.4mm] (earthquakeplot.east) to [out=0,in=240] (-4,0.9);
            \draw [-, color={rgb,255:red,72; green,154; blue,119}, line width=0.4mm] (earthquakeplot.east) to [out=0,in=0] (-4,0.29);
            \draw [-, color={rgb,255:red,72; green,154; blue,119}, line width=0.4mm] (earthquakeplot.east) to [out=0,in=120] (-4,-0.3);
            \draw [-, color={rgb,255:red,72; green,154; blue,119}, line width=0.4mm] (earthquakeplot.east) to [out=0,in=120] (-4,-1);
            \draw [-, color={rgb,255:red,72; green,154; blue,119}, line width=0.4mm] (earthquakeplot.east) to [out=0,in=120] (-4,-1.7);
            
            \node[rectangle,draw, minimum width=1.1cm, below=0.22cm of box.north] (PGA) {\footnotesize $Y_\textrm{pga}$};
            \node[rectangle,draw,below=0.1cm of PGA.south, minimum width=1.1cm] (PGV) {\footnotesize $Y_\textrm{pgv}$};
            \node[rectangle,draw,below=0.1cm of PGV.south, minimum width=1.1cm] (SA03) {\footnotesize $Y_\textrm{sa(.3s)}$};
            \node[rectangle,draw,below=0.1cm of SA03.south, minimum width=1.1cm] (SA1) {\footnotesize $Y_\textrm{sa(1s)}$};
            \node[rectangle,draw,below=0.1cm of SA1.south, minimum width=1.1cm] (SA3) {\footnotesize $Y_\textrm{sa(3s)}$};
            
            \draw [decorate,decoration = {brace},line width=0.2mm] (-2.9,2.88) --  (-0.2,2.88);
            \node (Test) at (-1.5,3.2) {\small $X$};
            
            \draw [decorate,decoration = {brace},line width=0.2mm] (0,2.88) --  (3.4,2.88);
            \node (Hidden) at (1.8,3.2) {\small Hidden};
            
            
            \draw [-, color={rgb,255:red,72; green,154; blue,119}, line width=0.4mm] (3.9,2.3) to [out=300,in=180] (box.west);
            \draw [-, color={rgb,255:red,72; green,154; blue,119}, line width=0.4mm] (3.9,1.6) to [out=300,in=180] (box.west);
            \draw [-, color={rgb,255:red,72; green,154; blue,119}, line width=0.4mm] (3.9,0.9) to [out=300,in=180] (box.west);
            
            \draw [->, color={rgb,255:red,72; green,154; blue,119}, line width=0.4mm] (3.9,0.29) to [out=0,in=180,shorten > = 10pt] (box.west);

            \draw [-, color={rgb,255:red,72; green,154; blue,119}, line width=0.4mm] (3.9,-0.3) to [out=60,in=180] (box.west);
            \draw [-, color={rgb,255:red,72; green,154; blue,119}, line width=0.4mm] (3.9,-1) to [out=60,in=180] (box.west);
            \draw [->, color={rgb,255:red,72; green,154; blue,119}, line width=0.4mm] (3.9,-1.7) to [out=60,in=180] (box.west);
            
            \node[above=0.001cm of box.north] (Hidden) {\small shape=($N,5$)};
            \end{tikzpicture}
    \caption{Overview of the task tackled in this paper. An example earthquake (P (green) and S(orange) wavefronts after 10s from the origin time) is shown on the left as red star. By taking the initial input length (10 s) of $X$, we predict the $Y$ values that characterize the earthquake at each node (representing a seismic station). $Y$ has to be inferred by exploiting waveform patterns in $X$, since $Y$ mostly reveals itself later on in the hidden part of the data (and do not necessarily occur at peaks). In addition, there could be \eg noise or sensor malfunctioning hindering information. We reduce the 10 s window length progressively by 1~s (each red block) at the time, to further complicate the task in Sect. \ref{sec:windowlenghts}.}
    \label{fig:regressionnew}
\end{figure*}

\subsection{Problem Definition}
The goal in this work is to regress various values from multivariate time series sensor data.
We test our models on seismic data, where the maximum intensity measurements of shaking at each station should be predicted.
We calculate values that are external to the input and do not depend necessarily on recent values, but rather on the whole length of the time series (see Fig. \ref{fig:regressionnew}).

Let $L$ be a symmetrically normalized laplacian matrix $L \in \mathbb{R}^{N\times N}$ where $N$ refers to the number of nodes in the graph, and $Z$ be a node feature matrix $Z \in \mathbb{R}^{2 \times N}$ that holds the latitude and longitude location of each node.
Given the input time-series $X \in \mathbb{R}^{E\times N\times T\times C}$ where $E$ is the number of earthquakes, $N$ the number of stations, $T$ the length of the time series and $C$ the amount of channels, our goal is to predict $Y \in \mathbb{R}^{5 \times N}$ , which refers to the 5 target parameters of the time series called PGV, PGA, SA(1s), SA(0.3s) and SA(3s) for each node in the graph. 
The task is iteratively complicated (see Fig. \ref{fig:regressionnew}) by reducing the input length $T$ given to $X$ in Sect. \ref{sec:windowlenghts}.

Our final regression problem can then be formulated as follows:

\begin{equation}
    f: L \times X \times Z \to Y
\end{equation}

\noindent where $f$ denotes the learning function, $L$ the graph, $X$ the time series input, $Z$ the node features and $Y$ the regression targets.

\subsection{Network creation}
Both undirected sensor networks were created by making use of the geographical locations of the seismic sensors. The adjacency matrix $A_{i,j}$ was calculated by taking all the pairwise geodesic (the shortest path between two points on a sphere) distances in km between each station (latitude, longitude), and taking the 1 - (min,max) scaled distance as the edge weight, since edges with a low distance should have a higher weight. Afterward, the resulting adjacency matrix can be filtered on the threshold $k$ to adjust the sparsity in the graph, \eg the higher the parameter $k$ is set, the fewer edges will retain in the graph ($k$ has a range between 0-1). Experimentation showed that in the CI network a threshold of 0.3 was most optimal, and 0.6 in the CW network (See Sect. \ref{sec:paramk}).

The adjacency matrix, however, still has to undergo some more changes to make it more suitable for GNNs (especially GCNs).
Therefore, we transform the adjacency matrix $A$ into the symmetrically normalized Laplacian matrix $L=I - D^{-1/2}AD^{-1/2}$ where $D$ refers to the Degree matrix containing the neighbors of each node and $I$ refers to the identity matrix of length $n$ nodes in a graph. 
A typical Laplacian would only consist of $L=D-A$, however, if nodes have a wide range of varying connectivity, vanishing gradient problems can occur \cite{kipf2016semi}. Therefore, the degree matrix is symmetrically normalized.
Lastly, the addition of the identity matrix helps with the GNN to also involve each node's own node features \cite{kipf2016semi}.

\begin{figure*}[htb]
    \begin{minipage}[t]{.48\textwidth}
        \centering
        \includegraphics[width=0.9\textwidth]{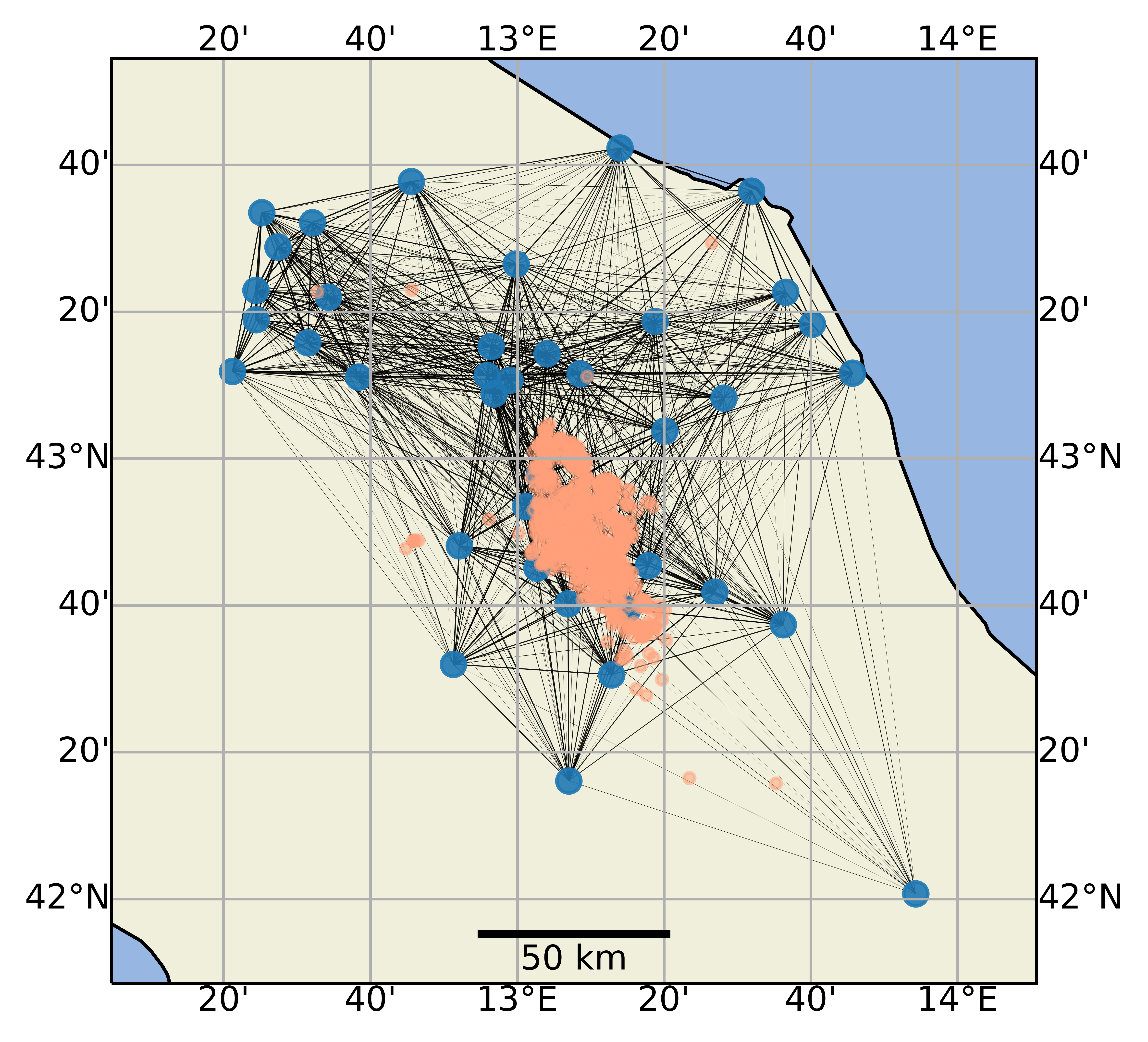}
        \subcaption{Network 1 (CI).}\label{fig:network1plot}
    \end{minipage}
    \hfill
    \begin{minipage}[t]{.48\textwidth}
        \centering
        \includegraphics[width=0.9\textwidth]{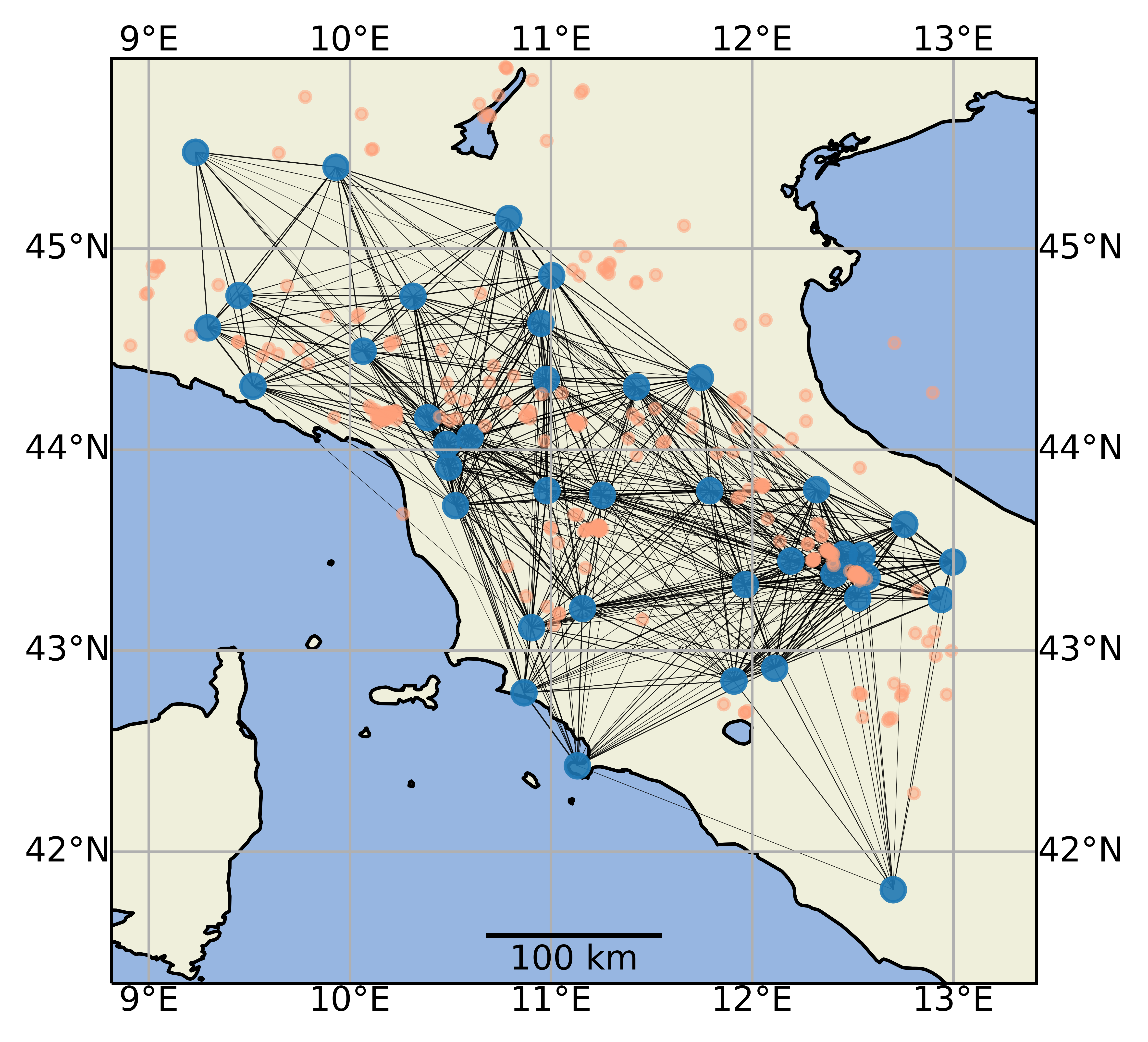}
        \subcaption{Network 2 (CW).}\label{fig:network2plot}
    \end{minipage}  
    \label{fig:network1and2plot_graph}
    \caption{Overview of the source-receiver geometries of the CI (a) and CW (b) seismic datasets.  The blue solid dots correspond to the seismic stations (nodes) and the orange dots refer to the earthquake epicenters. Notice the larger geographical area and the sparseness of the epicenters of the CW dataset when compared to CI (visible in the map scale at the bottom of both figures). The thickness of the lines connecting the stations (i.e., the edges) are inversely proportional to the distance of the connecting nodes as from the values of the adjacency matrix.}
    \label{fig:networkxplots}
\end{figure*}

 Fig.~\ref{fig:networkxplots} visualizes the resulting graphs; the nodes resulting from the seismic stations of the CI and CW datasets are shown in panels a) and b), respectively.
As mentioned before, looking at the geographical maps and the coordinates (latitude and longitude) on the axis of the plots, it is clear that the CW network covers a larger land area. 
In addition, the thickness of the edges in the figure is determined by the distance between two stations. 
The less distance between two stations, the higher the edge weight. 
A higher weight will help the GNN with determining which stations will most likely inhibit similar behavior in their sensor readings, improving the IMs' prediction.

\subsection{Abstracted framework}

\begin{figure}[htb]
    \centering
    \includegraphics[width=0.25\textwidth]{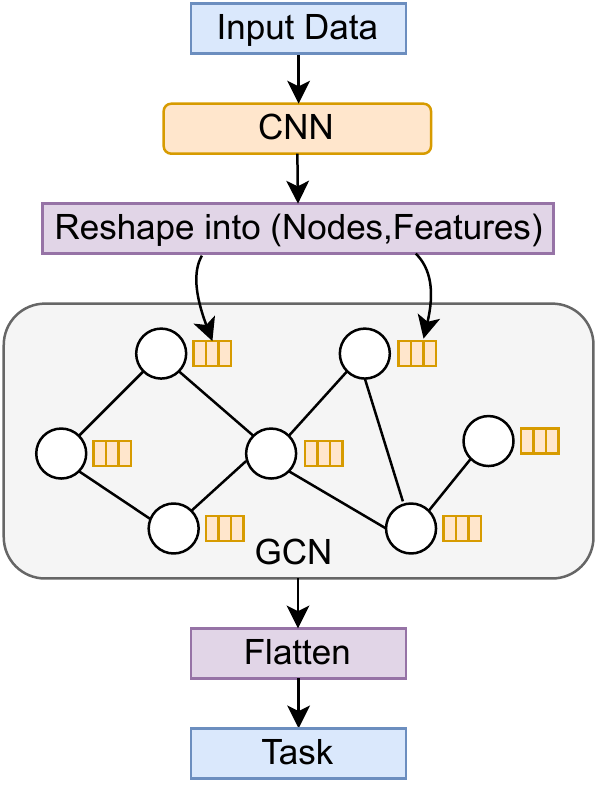}
    \caption{Abstract overview of our GNN implementation for multivariate time series processing.}
    \label{fig:gcn_block}
\end{figure}

Fig.~\ref{fig:gcn_block} presents an abstracted overview of the building blocks of our proposed architecture, which can therefore also be instantiated for other tasks, such as time series classification.
In summary, our proposed architecture of a GNN for time series regression (TISER-GCN) contains the following main contributions compared to previous work, as we will detail below:
\begin{enumerate}
    \item To obtain node features, we apply a 1D convolutional layer for feature extraction on the individual nodes using a wide kernel~\cite{app112311429,hoogen2020improvedWDCNN} on the input data as in \cite{jozinovic2020rapid}.
    \item To obtain the graph, the set of stations in the seismic area are considered as nodes, with the distance between them as edges.   
    \item A GNN (utilizing GCN layers from \cite{kipf2016semi}) of $n$ layers is implemented for processing these feature vectors calculated by the convolutional layers as node features. While in other GNN papers the node features each measure a unique aspect about a node (\eg the age and friend count in a social network), we demonstrate that GCNs can also learn from features that are sequential in time. 
    \item In our case, as described below, we focus on a \emph{regression task} on seismic data, focusing on predicting ground shaking at a set of seismic stations.
\end{enumerate}

\subsection{Model Implementation}\label{sec:modelarchitecture}
This section introduces the version of our abstracted implementation applied to a regression task on seismic data. For providing a complete picture of the model, the source code is available\footnote{https://github.com/StefanBloemheuvel/GCNTimeseriesRegression}.

\begin{figure}[t]
    \centering
    \includegraphics[width=0.45\textwidth]{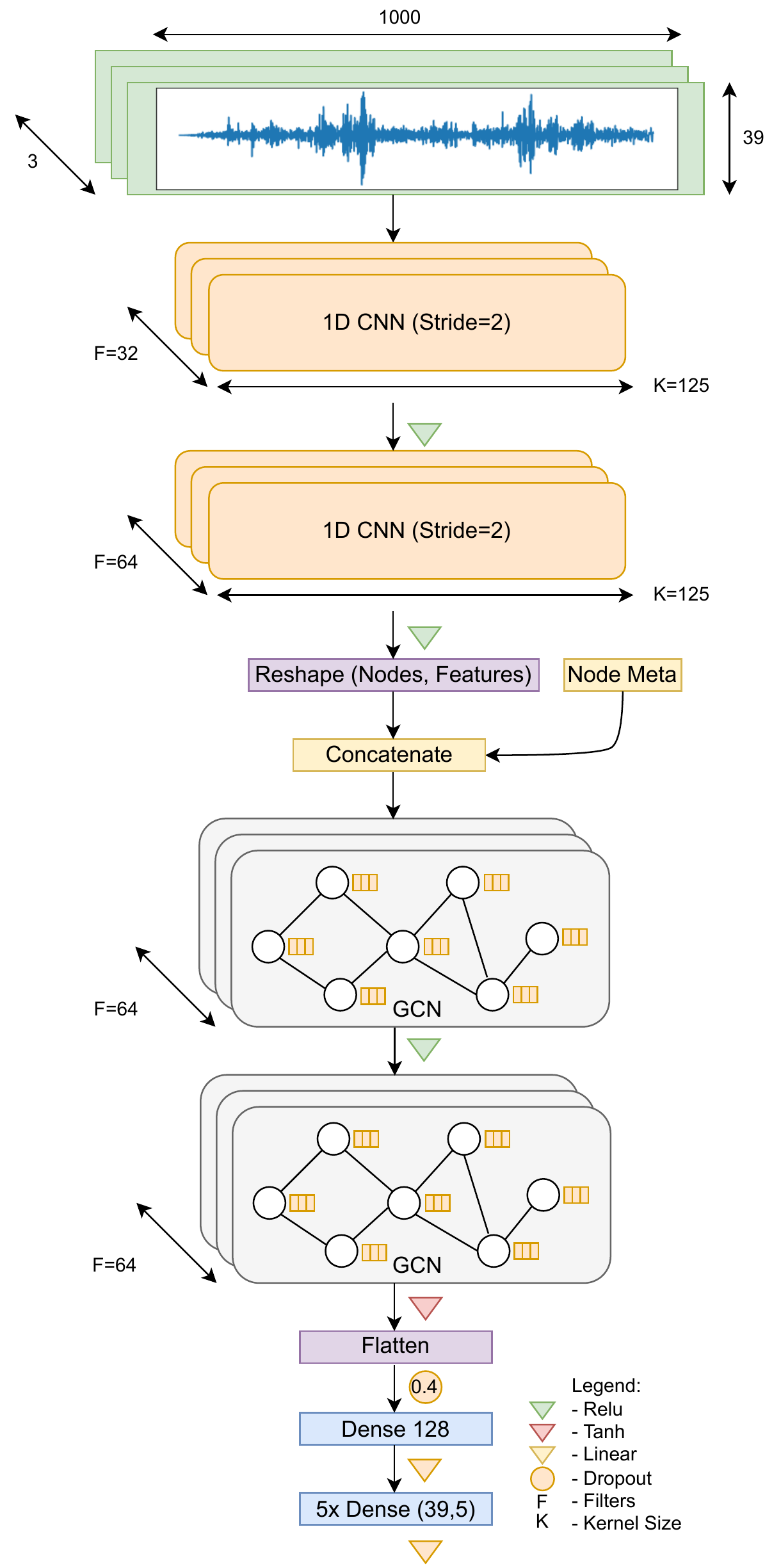}
    \caption{Overview of the proposed architecture. The features from two convolutional layers are used as node features in the GCNs. After the two GCN layers (which are used for inter-station related feature extraction), the data is flattened to retain as much information as possible. Afterward, the output is fed to fully connected layers.}
    \label{fig:modelarchitecture}
\end{figure}

The first block of our proposed model uses as input 10 s of 3-channel seismic waveform data sampled at 100 Hz, \ie a time series from each station of each earthquake. 
See Fig. \ref{fig:modelarchitecture} and Sect. \ref{sec:dataset} for a full overview of the model and the dataset. After that, convolutional, graph-convolutional and post-processing layers are applied.

\subsubsection{CNN for Feature Extraction}
In the second block of our model, two 1D convolutional layers act as feature extractors by using wide kernel sizes, small strides, increasing filters, kernel regularization and a ReLU activation function, which has proven to be useful for 1D time series data~\cite{jozinovic2020rapid,app112311429}.
The purpose of these convolutional layers is to learn the temporal patterns of each station.
Afterward, the output of the second convolutional layer, which has shape $(N, T, F)$ where $N$ refers to the number of nodes, $T$ the remaining length of the time series and $F$ the number of filters, is reshaped to make the dimensions fitted for the graph convolutional layers. These layers typically need an input of $(N,F)$ where $F$ now refers to a one-dimensional vector $[x_{1},x_{2}\dots x_{n}]$ for each node in the graph.
To this reshaped feature vector, features (latitude, longitude) of each node are added as node meta data.
Therefore, the feature vector of each node now consists of time series features from the convolutional layers and classical node features.
This addition of this node metadata has showed to improve performance in \cite{jozinovic2021transfer}.

\subsubsection{GCN processing}
Next follows the graph convolutional layers used from \cite{kipf2016semi}.
While in \cite{jozinovic2020rapid} the third convolutional layer gathers the cross-station information, here the graph convolutional layers takes this role, since they use the features from the convolutional layer as node features for each node.
More concretely, each node $N$ receives one of the feature vectors of dimension $(N,F)$ as node features where $F$ is the length of the feature vector (see Fig. \ref{fig:gcn_block}).
The two graph convolutional layers use these features of the nodes by reducing them to ($N$,64) by both containing 64 filters.
Considering the hyperparameters, experimentation revealed that starting with a ReLU activation function followed by a TanH works best.
In addition, bias was set to false (as suggested by \cite{kipf2016semi}) and the same kernel regularizer was used as in the convolutional layers.

\subsubsection{Postprocessing}
A common practice in the graph literature is using global graph pooling operators such as max or average-pooling \cite{defferrard2016convolutional,ying2018hierarchical,simonovsky2017dynamic}.
These pooling techniques take the embeddings of all the nodes in a graph and globally pool these together by an aggregation function (max, sum, mean ...)
However, this procedure reduces the feature vector from $(N,F)$ into a single vector $F$ regarding the number of nodes and filters used in the previous layer.
This means that the graph is reduced to essentially one node, which is not desirable in our task, because this is defined as a node-level regression task for all the nodes in the graph, in contrast to a graph-level task~\cite{ying2018hierarchical}.
Therefore, the output of the final graph convolutional layer is directly flattened and then fed to the fully connected layer.
The output of this layer is then given to five fully connected \emph{regression} layers.
These layers represent the regression target variables called PGV, PGA, SA(0.3 s), SA(1 s) and SA(3 s) for each of the nodes.

\subsection{Software and computer}
Python was used in combination with Tensorflow\footnote{https://www.tensorflow.org/} and Keras\footnote{https://keras.io/} to develop the proposed models. The GCN layer is derived from Spektral\footnote{https://graphneural.network/}.
Calculations are done with support of Numpy\footnote{https://numpy.org/} and table formatting with Pandas\footnote{https://pandas.pydata.org/}. 
Furthermore, to reduce the overall training time, the models are trained on a dedicated server with two Intel Xeon CPUs (3.2 GHz), 256 GB RAM and a Nvidia Quadro RTX3956000 (24 GB) GPU. After training, the models are fairly small (around 6 MB) and are deployable, \eg on standard PC hardware as well as edge computing platforms.

\subsection{Model training}
Regarding model training, 80\% of each dataset was used for training and 20\% for testing.
The train set was then randomly split by k-fold cross validation with $k=5$. 
The average MAE, MSE and RMSE scores on the test set were taken as the results.
This entire procedure was repeated 5 times with different seeds, which generates 5 distinct train-test splits to generalize the results (i.e., the large amplitudes from the less frequent large earthquakes in a test set can influence the scores if more of them are assigned to a test set). 

The model used a batch size of 20 and 100 training epochs with early stopping - patience of 10.
The same optimizer as in our baseline \cite{jozinovic2020rapid} was used, \ie RMSprop with mostly standard settings~\cite{hinton2012neural}.
Lastly, MSE (Mean Squared Error) was used as the loss function when training the models:

\begin{equation}
\mathrm{MSE}=\frac{1}{n} \sum_{i=1}^{n}\left(y_{i}-\hat{y}_{i}\right)^{2},
\end{equation}

\noindent where $y_i$ are the actual values and $\hat{y_i}$ the predictions, since it penalizes larger errors more than \eg Mean Absolute Error

\subsection{Baseline models}
We compare our work to the model presented by \cite{jozinovic2020rapid}. This baseline model was given the exact same input data except for the graph, \ie both models received the time series per station for every earthquake combined with the latitude and longitude node features of every station, as proven to be effective in \cite{jozinovic2021transfer}. 

In addition, we also examine the performance of GAT \cite{velickovic2018graph} layers and an adjusted version of \cite{kim2021graph} for our task. The GAT layers are set with 8 channels and 8 attention heads in order to match the number of parameters of our model, all other hyperparameters were unchanged. To adapt the model of \cite{kim2021graph} to our task, the last layers used for classification were altered for regression, a weighted initial adjacency matrix was supplied, and node features were added.

\begin{table*}[htb]
    \small
    \centering
    \caption{Overview of possible parameter settings used for the grid search optimization of the ML models.}
    \begin{tabular}{ccc}
    \toprule
        Model & Parameter & Option range  \\
        \midrule
         \textbf{K-NN} & K                  & 1-20 (1 per step) \\
                       & Weight options     & Uniform or distance \\
        \midrule
        \textbf{SVM}   & C                  & 10-40 (5 per step) \\
                       & Gamma              & [0.0001,0.001] \\
                       & Kernel             & Linear or radial basis function \\
        \midrule
        \textbf{XGBoost}   & $N$ estimators                  & 100-1000 (100 per step) \\
               & Max depth              & 5-15 (5 per step) \\
               & Gamma             & [0.0,0.1,0.2,0.3,0.4] \\
        \midrule
        \textbf{RF}    & $N$ estimators   & 100-1000 (100 per step) \\
                       & Max features & square root or Log2 \\
        \bottomrule
    \end{tabular}
    \label{tab:ml_options}
\end{table*}

Furthermore, our proposed model is also compared with traditional machine learning (ML) algorithms: k-nearest neighbors (K-NN), extreme gradient boosting (XGBoost), random forest (RF) and support vector machines (SVM). Since these models are not designed to process multidimensional data, features were calculated from both the time and frequency domain. These features are derived from several studies \cite{mazilu2013feature,masiala2019feature}. 
From the time domain, the mean, standard deviation, variance, median, minimum, maximum and range (maximum-minimum) are used.
From the frequency-domain, specifically the signal energy $E =\sum(\operatorname{fft}x_{i})^2$ and signal power $P =\sum\frac{(\operatorname{fft}x_{i})^2}{\sum_{i}^{t}}$ were used.
For each of the ML models, grid search optimization in combination with fivefold cross-validation was used to assess which models performed best. 
Table \ref{tab:ml_options} describes all the options of the grid search optimization.

Lastly, we also compare the best performing deep learning models for each network without node features added, to examine their effect on the performance.
These results are visible in the Ablation study in Sect. \ref{sec:ablation}.
Here, the impact of the spatial information can be observed.

\section{Results}\label{sec:results}
In this section, the results from the multivariate regression task are shown. In addition, our model is tested on different input window lengths to further complicate the task.

\subsection{IM prediction}\label{sec:impredictions}
\begin{figure*}[htb]
    \centering
    \begin{minipage}{0.27\textwidth}
        \includegraphics[width = \textwidth]{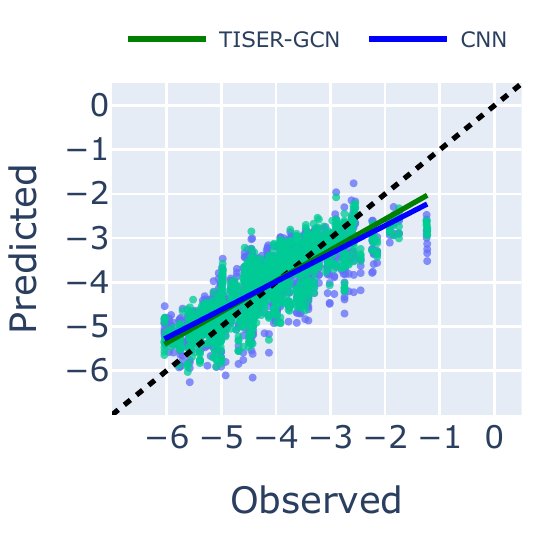}
        \caption*{(a) \small{PGV}}
    \end{minipage} 
    \begin{minipage}{0.27\textwidth}
        \includegraphics[width = \textwidth]{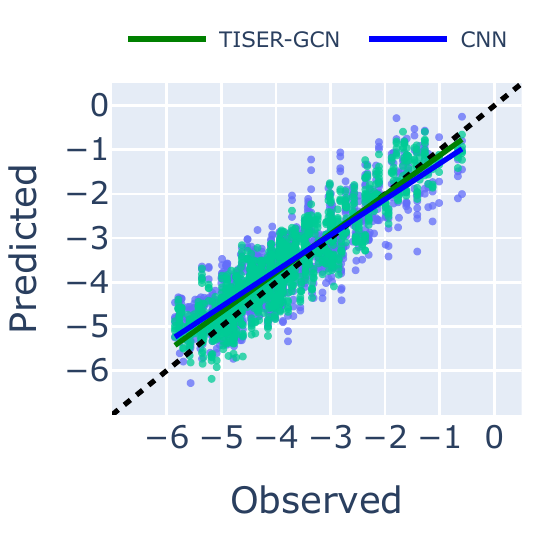}
        \caption*{(b) \small{PGA}}
    \end{minipage} 
    \begin{minipage}{0.27\textwidth}
        \includegraphics[width = \textwidth]{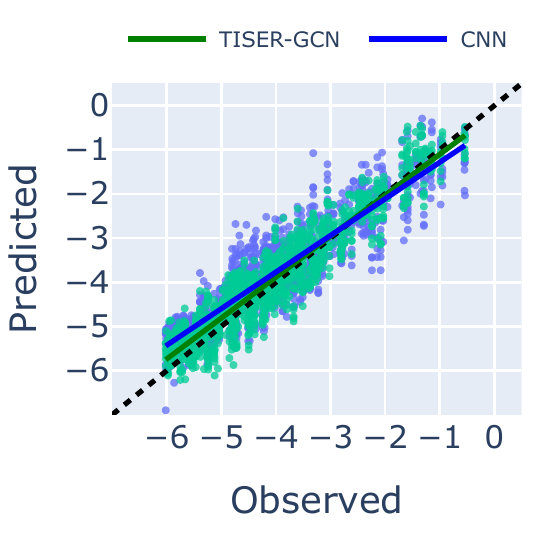}
        \caption*{(c) \small{SA (0.3 s)}}
    \end{minipage} 
    \begin{minipage}{0.27\textwidth}
        \includegraphics[width = \textwidth]{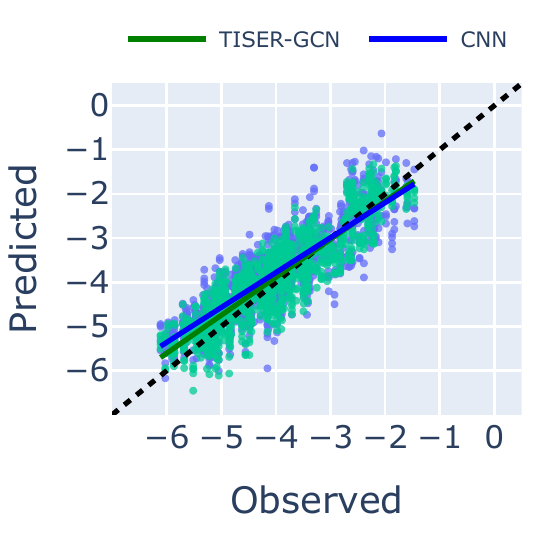}
        \caption*{(d) \small{SA (1 s)}}
    \end{minipage} 
    \begin{minipage}{0.27\textwidth}
        \includegraphics[width = \textwidth]{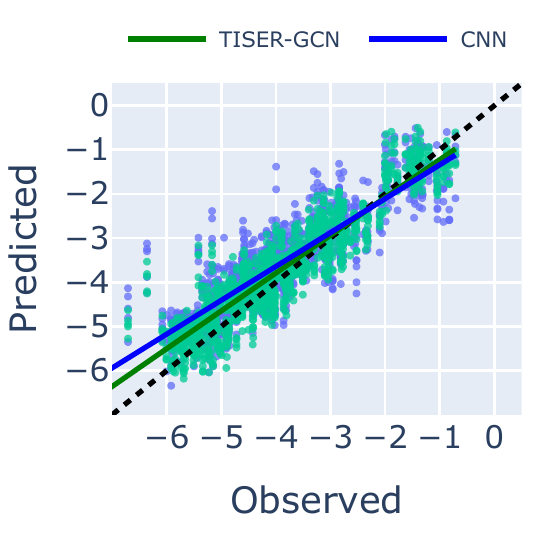}
        \caption*{(e) \small{SA (3 s)}}
    \end{minipage}
    \caption{Residual plots of the predicted against the true 5 IMs of the CW dataset (displayed in logarithmic ($\log_{10}$) form). The blue line and points display the results of~\cite{jozinovic2020rapid} CNN, the green line and points of TISER-GCN. The dotted black line resembles a perfect prediction score. The units are $m/s$ for PGV and $m/s^{2}$ for PGA and SA.}
    \label{fig:scatterplots}
\end{figure*}

\begin{figure*}[htb]
    \begin{minipage}[t]{.49\textwidth}
        \centering
        \includegraphics[width=0.83\textwidth]{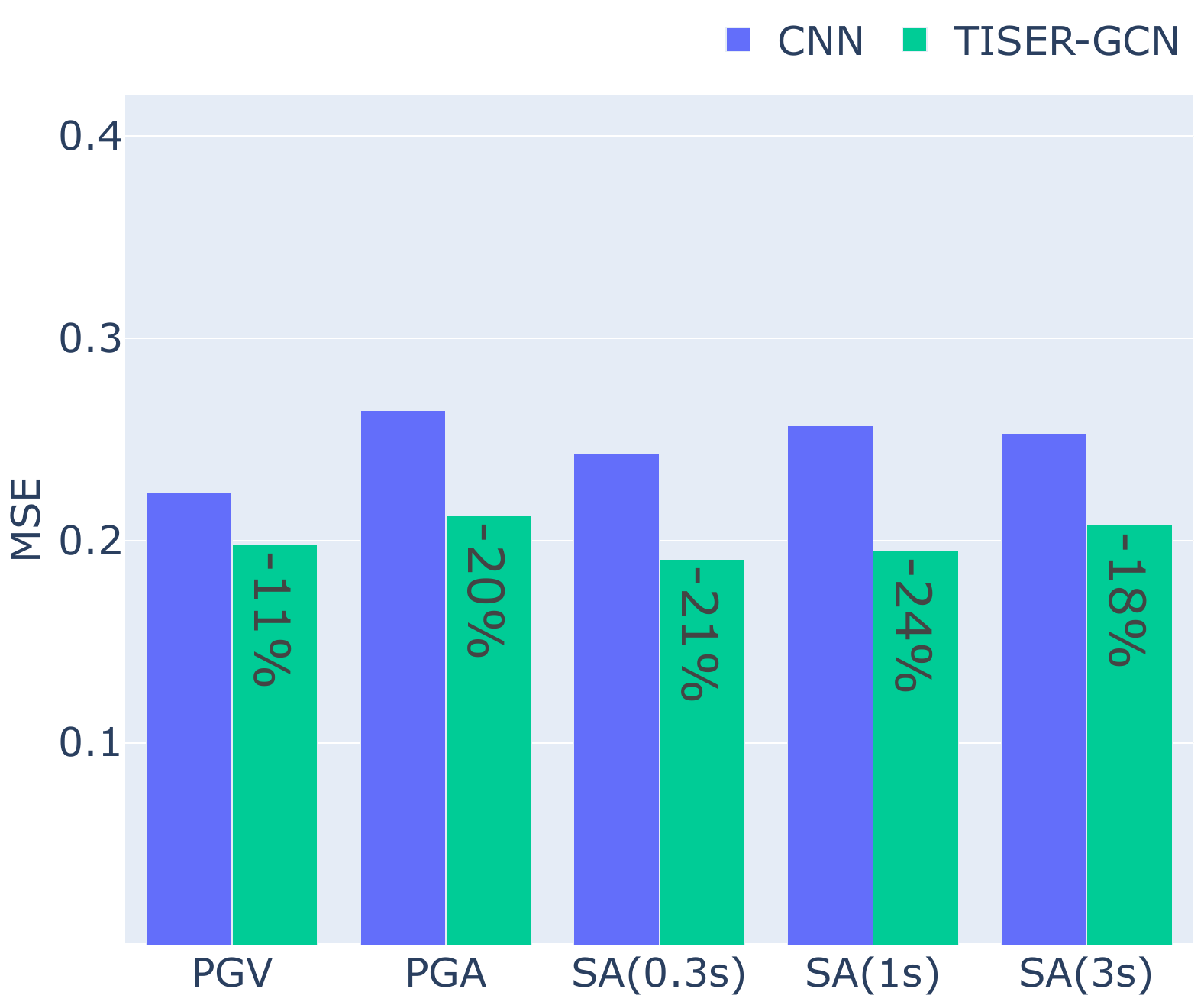}
        \subcaption{Network 1 (CI).}\label{fig:msenetwork1plot}
    \end{minipage}
    \hfill
    \begin{minipage}[t]{.49\textwidth}
        \centering
        \includegraphics[width=0.83\textwidth]{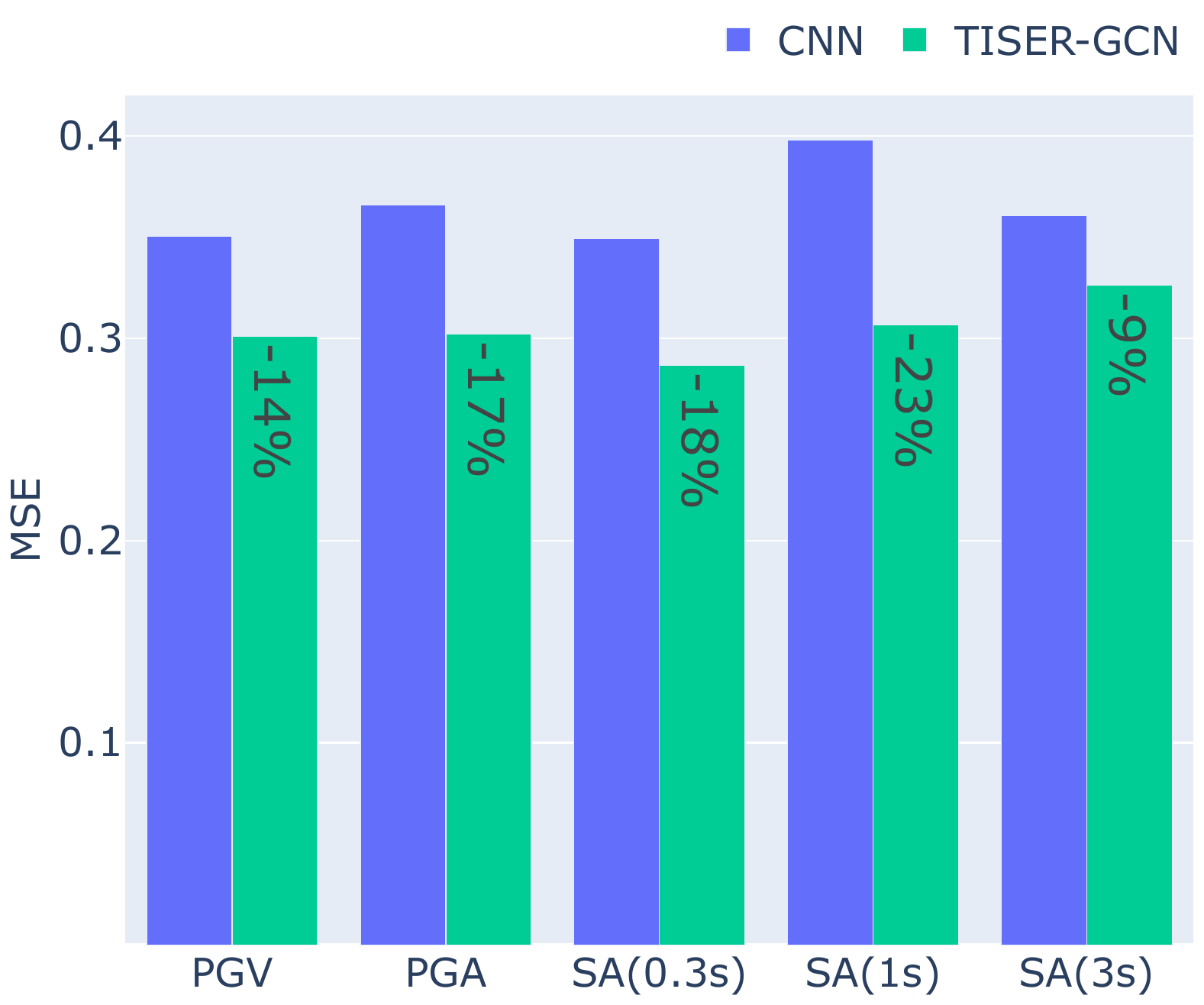}
        \subcaption{Network 2 (CW).}\label{fig:msenetwork2plot}
    \end{minipage}  
    \caption{MSE scores for each metric where the blue bar represents the original CNN model by \cite{jozinovic2020rapid} (our initial baseline) and the green bars our proposed model.}
    \label{fig:msenetwork1and2plot}
\end{figure*}

The results of the IM prediction are visible in Table \ref{table:individualresults} and Fig. \ref{fig:scatterplots} and \ref{fig:msenetwork1and2plot}. All the algorithms perform better on the CI network than on the CW network. 
Such behavior is expected since the CW dataset contains fewer earthquakes (266 against 915) than the CI dataset and, in contrast with the CI, their spatial distribution is sparse. 
In addition, the CW network covers a larger area, with greater distances between the stations, a larger depth range of the hypocenters and a larger variability in the geological settings.
As an extra test, 266 samples were taken from the 915 earthquakes in the CI dataset to mimic the conditions of the CW dataset while preserving the densely located earthquakes characteristic of the CI dataset.
In general, this resulted in an increase in MSE of ~30\% for our model (TISER-GCN) and ~42\% for the CNN model from \cite{jozinovic2020rapid}, showing the importance of having enough samples for learning.

\begin{table*}
\centering
\small
\caption{Individual results (best performing in bold) of each IM metric for the support-vector-machine (SVM), K-nearest-neighbors (KNN), XGBoost, random forest (RF), the CNN model from \cite{jozinovic2020rapid}, GAT-layers, adjusted implementation of \cite{kim2021graph} and our proposed model (TISER-GCN).}
\scalebox{0.743}{
\label{table:individualresults}
\begin{tabular}{lccc|ccc|ccc|ccc|ccc}
\toprule
  & \multicolumn{3}{c}{PGA} & \multicolumn{3}{c}{PGV} & \multicolumn{3}{c}{PSA03} & \multicolumn{3}{c}{PSA1} & \multicolumn{3}{c}{PSA3} \\
  & MAE & MSE & RMSE & MAE & MSE & RMSE & MAE & MSE & RMSE & MAE & MSE & RMSE & MAE & MSE & RMSE \\
\midrule
\textbf{CI network} &  &  &  &  &  &  &  & &   & &  &  &  & \\
 SVM & 0.43 & 0.36 & 0.60 & 0.47 & 0.43 & 0.65 & 0.47 & 0.41 & 0.64 & 0.44 & 0.37 & 0.61 & 0.45 & 0.40 & 0.63 \\
 KNN & 0.41 & 0.32 & 0.56 & 0.44 & 0.37 & 0.61 & 0.45 & 0.37 & 0.61 & 0.43 & 0.35 & 0.59 & 0.44 & 0.38 & 0.62 \\
 XGBoost & 0.38 & 0.28 & 0.53 & 0.41 & 0.32 & 0.57 & 0.42 & 0.33 & 0.57 & 0.41 & 0.31 & 0.56 & 0.41 & 0.33 & 0.58 \\
 RF & 0.38 & 0.28 & 0.53 & 0.41 & 0.32 & 0.57 & 0.42 & 0.33 & 0.57 & 0.41 & 0.31 & 0.56 & 0.41 & 0.33 & 0.57 \\
 GAT & 0.39 & 0.30 & 0.54 & 0.36 & 0.26 & 0.49 & 0.36 & 0.26 & 0.49 & 0.38 & 0.28 & 0.52 & 0.37 & 0.28 & 0.52 \\
 Jozinovic et al. \cite{jozinovic2020rapid} & 0.34 & 0.22 & 0.46 & 0.35 & 0.26 & 0.50 & 0.36 & 0.24 & 0.48 & 0.35 & 0.26 & 0.49 & 0.36 & 0.25 & 0.49 \\
 Kim et al. \cite{kim2021graph} & 0.35 & 0.26 & 0.49 & 0.33 & 0.23 & 0.47 & 0.33 & 0.23 & 0.47 & 0.33 & 0.24 & 0.48 & 0.33 & 0.24 & 0.48 \\
 TISER-GCN & \textbf{0.31} & \textbf{0.20} & \textbf{0.44} & \textbf{0.32} & \textbf{0.21} & \textbf{0.45} & \textbf{0.31} & \textbf{0.19} & \textbf{0.43} & \textbf{0.31} & \textbf{0.20} & \textbf{0.43} & \textbf{0.32} & \textbf{0.21} & \textbf{0.45} \\
\cline{1-16}
 \textbf{CW network} &  &  &  &  &  &  &  & &   & &  &  &  & \\
 GAT & 0.54 & 0.49 & 0.68 & 0.56 & 0.52 & 0.70 & 0.55 & 0.52 & 0.70 & 0.53 & 0.49 & 0.68 & 0.58 & 0.56 & 0.72 \\
 SVM & 0.51 & 0.43 & 0.66 & 0.56 & 0.51 & 0.71 & 0.60 & 0.58 & 0.77 & 0.56 & 0.51 & 0.72 & 0.47 & 0.40 & 0.63 \\
 KNN & 0.52 & 0.45 & 0.67 & 0.57 & 0.51 & 0.71 & 0.61 & 0.60 & 0.78 & 0.57 & 0.53 & 0.73 & 0.48 & 0.41 & 0.64 \\
 XGBoost & 0.50 & 0.42 & 0.65 & 0.54 & 0.48 & 0.69 & 0.59 & 0.57 & 0.75 & 0.55 & 0.51 & 0.72 & 0.46 & 0.39 & 0.62 \\
 RF & 0.49 & 0.40 & 0.63 & 0.54 & 0.47 & 0.68 & 0.58 & 0.56 & 0.75 & 0.55 & 0.50 & 0.71 & 0.46 & 0.39 & 0.62 \\
 Kim et al. \cite{kim2021graph} & 0.45 & 0.35 & 0.59 & 0.48 & 0.40 & 0.62 & 0.46 & 0.38 & 0.60 & 0.45 & 0.35 & 0.58 & 0.46 & 0.37 & 0.60 \\
 Jozinovic et al. \cite{jozinovic2020rapid} & 0.44 & 0.35 & 0.58 & 0.46 & 0.37 & 0.59 & 0.44 & 0.35 & 0.58 & 0.48 & 0.40 & 0.62 & 0.45 & 0.36 & 0.58 \\
 TISER-GCN & \textbf{0.41} & \textbf{0.30} & \textbf{0.54} & \textbf{0.41} & \textbf{0.30} & \textbf{0.54} & \textbf{0.40} & \textbf{0.29} & \textbf{0.52} & \textbf{0.42} & \textbf{0.31} & \textbf{0.54} & \textbf{0.43} & \textbf{0.33} & \textbf{0.56} \\
\bottomrule
\end{tabular}
}
\end{table*}

When examining the individual performance of the models, TISER-GCN outperforms the best performing baselines (the model from \cite{kim2021graph} on the CI network and the CNN model from \cite{jozinovic2020rapid} on the CW network, respectively) by a large margin on each of the five metrics of ground motion. 
Especially in the PGA and SA(1 s) metrics this performance gain is visible.
Our model improves an average of 7\% on MAE, 16.1\% on MSE and 8.3\% on RMSE compared to the best baseline for the CI network. Considering the CW network, an improvement of 9.1\% on MAE, 16.5\% on MSE and 8.4\% on RMSE is achieved.

Lastly, it is interesting to see the relatively weak performance of the GAT-based model. A possible explanation could be that the explicitly defined spatial information in the graphs (the distances between the stations) is crucial to make sense of the time series data, which the GAT-layers infer themselves via self-attention.
Therefore, perhaps the time series features are too complex for the GAT-layers to learn such representations, especially in the CW network.

Considering the results of the models on each individual earthquake and each IM metric, Fig. \ref{fig:scatterplots} shows the observed versus the predicted IM values of the CW network for both \cite{jozinovic2020rapid} and TISER-GCN as residual plots.
The blue lines (and residuals) reveal the performance of the CNN model \cite{jozinovic2020rapid} and the green lines (and residuals) TISER-GCN. 
The lines were calculated with an ordinary least squares to better visualize the difference in prediction bias between the two models. 
We observe, that better performance of our model comes also in slight reduction of the bias for large IM values (less underestimation), which is also of great value for the seismological applications (for more information, see the original CNN model~\cite{jozinovic2020rapid}).

\begin{figure}[htb]
    \centering
    \includegraphics[width=0.475\textwidth]{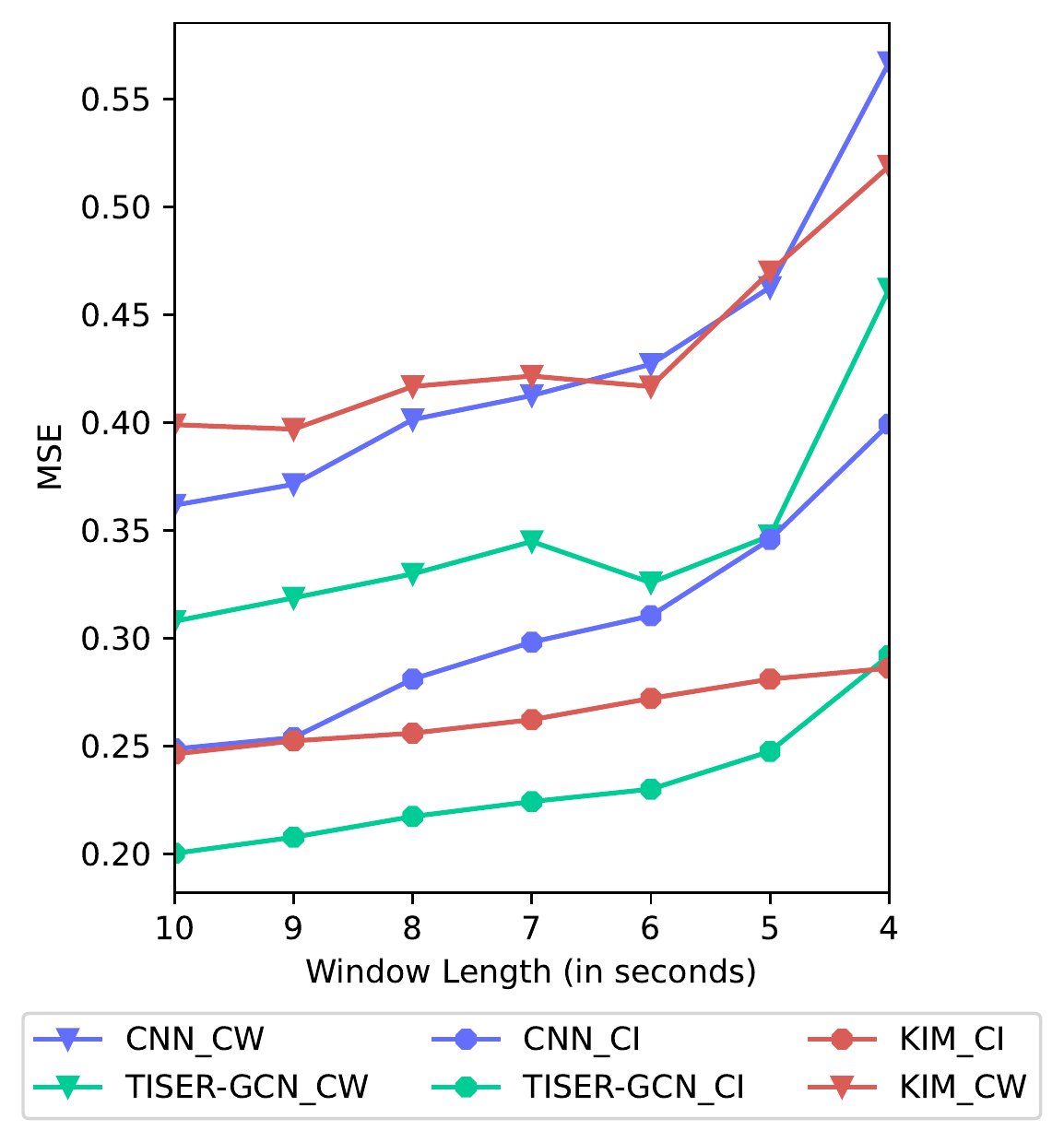}
    \caption{MSE at different input window lengths for each model in both networks. Read the figure from left (initial 10 s window from Sect. \ref{sec:impredictions}) to right. Our model is capable of achieving approximately the same MSE scores when given half the input, highlighting the power of GCNs with processing spatial information.}
    \label{fig:ablation_windowlength}
\end{figure}

Lastly, the characteristics of almost all the deep learning models are highly similar.
The models do not deviate concerning Ms/step per epoch. 
However, there is a small decrease in the number of parameters, since our TISER-GCN model has 6.7\% fewer parameters than the CNN model from \cite{jozinovic2020rapid}.
The GAT-model has an equal amount of parameters as our model, and the adapted GCN model from \cite{kim2021graph} originally already has fewer parameters.

\subsubsection{Variation in different window lengths}\label{sec:windowlenghts}
Sect. \ref{sec:impredictions} showed the results based on a 10 s input window length.
However, it is interesting to investigate the results of both the CNN and the best performing GNN-based models when this window length is reduced, since smaller window lengths could translate in earlier responses.
However, a smaller input length creates a more difficult setting, since fewer stations have received enough information to predict the IMs at further away stations.
Therefore, all the window lengths between 10 s and 4 seconds (it is not reasonable to reduce the input window further with this specific application) and their corresponding MSE scores are displayed in Fig.~\ref{fig:ablation_windowlength}. 

We find that we can half the entire input window, while still achieve similar performance as the best performing baselines.
This reduction also results in a reduction of the model parameter size of 1.26 million to around 700 thousand (-44\%).

Overall, these results highlight the power of the GCN layers and our implementation.
Since GCN layers are designed to perform node feature sharing in its convolution procedure, it is still possible to transfer plenty of information between the nodes in a situation where half the input is provided. 
In the context of analyzing streaming time series data, these benefits are crucial \cite{domingos2001catching} since requiring less input translates to earlier responses.

\subsection{Experimentation: ablation study}\label{sec:ablation}
\subsubsection{Tuning hyperparameter k.}\label{sec:paramk}
The graph creation algorithm mentioned in Sect. \ref{sec:method}, used the hyperparameter $k$ to cutoff connections between stations that are too far away. 
$k$ falls between 0-1, where 0 means that no edges are filtered, and 1 means that all edges are filtered.
Table \ref{table:ablation} presents the results of our model when tuning this parameter on both networks and at what distance the cutoff will be.
If the MSE scores of two values of $k$ would be highly similar, one would choose a higher $k$ over a lower $k$, due to the computational advantages of sparser graphs.

While in the CI network the results for $k$ are more promising with lower values of $k$, different results are visible in the CW network.
An explanation can be found in the very different characteristics (see Sect. \ref{sec:dataset}) of the two networks.
That is, a too low cutoff has bigger effect applied to a widely spaced network (\ie CW) than on a more concentrated network (\ie CI).
In practice, having more edges than necessary in a largely spread network (CW network) apparently confuses the GCN layers in this experiment, perhaps since stations in less dense networks show different behavior sooner than in more densely networks (\eg the CI network).
These results highlight the importance of this preprocessing step when designing graphs.
However, it is an easily interpretable hyperparameter that once determined can help GCNs achieve great performance.

\begin{table*}[htb]
\centering
\small
\caption{Ablation results of the minimum distance cutoff hyperparameter $k$ for both networks.}
\begin{tabular}{@{}lccccr@{}}
\toprule
Network  & $k$          &  Cutoff (km)    & Edges        &  Avg. Degree Centrality & MSE            \\ \midrule
CI       & 0.6          & 92              & 497          &  0.67                   & 0.224          \\
         & 0.5          & 115             & 609          &  0.82                   & 0.209         \\
         & 0.4          & 138             & 687          &  0.92                   & 0.191          \\
         & 0.3          & 159             & 722          &  0.97                   & 0.189          \\
         & 0.2          & 177             & 729          &  0.98                   & 0.188          \\ 
         & 0.1          & 204             & 733          &  0.99                   & 0.191          \\ \midrule
         
CW       & 0.6          & 217             & 493          &  0.67                   & 0.307 \\
         & 0.5          & 271             & 601          &  0.81                   & 0.314          \\
         & 0.4          & 325             & 660          &  0.89                   & 0.310          \\
         & 0.3          & 377             & 705          &  0.95                   & 0.322          \\
         & 0.2          & 429             & 731          &  0.99                   & 0.360          \\
         & 0.1          & 485             & 739          &  0.99                   & 0.368          \\
\bottomrule
\end{tabular}
\label{table:ablation}
\end{table*}

\subsubsection{Effect of Node Metadata}
The impact of the node features can be examined by inspecting the results if no features (latitude, longitude) were supplied. 
Without node features, which have proven to be crucial in our difficult prediction task \cite{jozinovic2021transfer}, the CNN from \cite{jozinovic2020rapid} and TISER-GCN score approximately equal.
Both have an MSE of 0.26 on the CI network, and the GCN from \cite{kim2021graph} achieves a promising 0.24 MSE.
On the CW network, \cite{jozinovic2020rapid} scores 0.50, TISER-GCN 0.51, and the GCN from \cite{kim2021graph} scores a 0.37.
These scores show how difficult this task is without including the spatial information from the stations, which shows to be crucial for this task, confirming the insights from~\cite{jozinovic2021transfer}. 
In addition, they show the promising results from \cite{kim2021graph} model with no metadata added.

However, once the features are added, changes in performance become visible.
The model from \cite{jozinovic2020rapid} improves 5\% on the CI network and 26\% on the CW network, whereas our model improves 24\% on the CI network and 41\% on the CW network.
In contrast, the GCN from \cite{kim2021graph} does not improve at all, staying at 0.24 and 0.37 MSE, respectively.
Therefore, our TISER-GCN appears to be more capable of using this spatial information to improve the learned feature representation, by combining the graph information with the time series data while exceptionally well exploiting the information contained in the spatial metadata in the graph convolutional layers.
However, we want to emphasize that the model from \cite{kim2021graph} was not originally optimized for this task, but for earthquake classification.

\section{Conclusions}\label{sec:conclusions}
In this work, the use of multivariate time series regression with graph neural networks was presented. 
Our method (TISER-GCN) proposed a unique way to leverage features from convolutional layers as node features in a GCN.
The proposed model is tested on two seismic datasets with different characteristics, demonstrating the generalizability of the model.
Our model outperforms the best performing baselines by 16.1\% on average on the CI dataset and 16.5\% on the CW dataset in terms of MSE.
Therefore, besides the original baseline of \cite{jozinovic2020rapid}, the adjusted version of \cite{kim2021graph} and GAT layers \cite{velickovic2018graph} also showed weaker performance compared to our model.
The experiments demonstrate the impressive power of TISER-GCN, \ie of our proposed GCN model when processing spatial information, since the tested deep learning models were provided with the same input.
More crucially, especially when taking into account the use case of early warning systems, our model can match the performance of the baselines by using half of the input window length on both datasets. 
Such a reduction can help with faster earthquake early warnings (half the input means a faster response).

One important message which we want to emphasize is that the architecture of the original CNN model by \cite{jozinovic2020rapid} could also be improved from an only CNN perspective.
However, to make the comparison more fair, and to observe more directly the actual effect of the GCN layers on the prediction results, we decided to alter the model architecture as little as possible to demonstrate the difference in performance between a classical CNN approach compared to our proposed model.

For future research, other methods of creating the initial adjacency matrix will be investigated, because the results of the cutoff parameter experiments in the ablation study reveal the effect the graph creation steps have on the performance in the CW network.
Examples include; to keep the top-k edges for each node or using exponential decaying functions as in \cite{shuman2013emerging}.
Furthermore, other node features could be added to the node feature vector to improve predictions. For example, the angle between two stations could be added as an edge feature, or the absolute distance between two stations can be used. Here, also explanation techniques~\cite{luo2020parameterized,DBLP:conf/dsaa/SchwenkeA21} could yield interesting insights, in order to lead feature construction and modeling.

In addition, we plan to test our model on other (types of) datasets.
For example, both networks now consisted of 39 nodes.
It would be interesting how our architecture would scale to datasets featuring 200 or more nodes. 
Also, to test the transfer learning capabilities of our model, perhaps adding the spatial information to a pre-trained model on one dataset could make it more adaptable for other networks with different data characteristics.

Lastly, our architecture was built to be easily adaptable for other tasks (involving regression or classification).
Inspiration can be taken from the examples of \cite{tan2021time}, however, there could be as well many other types of datasets where scalar or vector value quantities, either associated to a single node or to the entire graph are to be predicted.  
Therefore, we encourage readers to take Fig. \ref{fig:gcn_block} as a departure point for other analysis tasks.

\section*{Declarations}
\subsubsection*{Funding}
This work has been funded by the Interreg North-West Europe program (Interreg NWE), project Di-Plast - Digital Circular Economy for the Plastics Industry (NWE729).

This work was also partially supported by the project INGV Pianeta Dinamico 2021 Tema 8 SOME (CUP D53J1900017001) funded by Italian Ministry of University and Research “Fondo finalizzato al rilancio degli investimenti delle amministrazioni centrali dello Stato e allo sviluppo del Paese, legge 145/2018.

\subsubsection*{Conflicts of Interest/Competing interests}
Not Applicable
\subsubsection*{Ethics Approval}
Not Applicable
\subsubsection*{Consent to participate}
Not Applicable
\subsubsection*{Consent for publication}
Not Applicable
\subsubsection*{Availability of data and material}
Data are available at \cite{jozinovic_dario_2020_3669969} (CI dataset) and \cite{dario_jozinovic_2021_5541083} (CW dataset).

\subsubsection*{Code availability}
There is a Github \href{https://github.com/StefanBloemheuvel/GCNTimeseriesRegression}{page} available with the corresponding code.

\subsubsection*{Authors' contributions}
SB and MA conceived of the idea and study, as well as the interpretation of the data, which was performed in a synergistic way together with DJ and AM. SB, MA and JH drafted the manuscript. All authors edited the manuscript. SB implemented the methods and algorithms supported by JH, and ran the experiments, in close collaboration with DJ and AM. DJ and AM provided the baseline model and the data. All authors read, reviewed and approved the final manuscript.

\renewcommand*{\bibfont}{\small}

\end{document}